\documentclass[conference]{IEEEtran}
\IEEEoverridecommandlockouts
\usepackage{cite}
\usepackage{amsmath,amssymb,amsfonts}
\usepackage{algorithm}
\usepackage{algpseudocode}
\usepackage{booktabs}
\usepackage{graphicx}
\usepackage{textcomp}
\usepackage{xcolor}
\usepackage{color}
\usepackage{siunitx}
\usepackage{multicol}
\usepackage{multirow}
\usepackage{bm}
\usepackage{subcaption}
\usepackage{url}
\usepackage{autobreak}
\usepackage[para]{footmisc}
\def\BibTeX{{\rm B\kern-.05em{\sc i\kern-.025em b}\kern-.08em
    T\kern-.1667em\lower.7ex\hbox{E}\kern-.125emX}}
\begin{document}

\title{Energy and Quality of Surrogate-Assisted Search Algorithms: a First Analysis
\thanks{This research is partially funded by the PID 2020-116727RB-I00 (HUmove) funded by MCIN/AEI/ 10.13039/501100011033; TAILOR ICT-48 Network (No 952215) funded by EU Horizon 2020 research and innovation programme; and Japan Society for the Promotion of Science 21K17826.}
}

\author{\IEEEauthorblockN{Anonymous Authors}}
\author{\IEEEauthorblockN{Tomohiro Harada}
\IEEEauthorblockA{\textit{Graduate School of Science and Engineering} \\
\textit{Saitama University}\\
Saitama, Japan \\
tharada@mail.saitama-u.ac.jp}
\and
\IEEEauthorblockN{Enrique Alba}
\IEEEauthorblockA{\textit{ITIS Software} \\
\textit{University of M\'{a}laga}\\
M\'{a}laga, Spain \\
eat@lcc.uma.es}
\and
\IEEEauthorblockN{Gabriel Luque}
\IEEEauthorblockA{\textit{ITIS Software} \\
\textit{University of M\'{a}laga}\\
M\'{a}laga, Spain \\
gabriel@lcc.uma.es}
}

\maketitle

\begin{abstract}
Solving complex real problems often demands advanced algorithms, and then continuous improvements in the internal operations of a search technique are needed. Hybrid algorithms, parallel techniques, theoretical advances, and much more are needed to transform a general search algorithm into an efficient, useful one in practice. In this paper, we study how surrogates are helping metaheuristics from an important and understudied point of view: their energy profile. Even if surrogates are a great idea for substituting a time-demanding complex fitness function, the energy profile, general efficiency, and accuracy of the resulting surrogate-assisted metaheuristic still need considerable research. In this work, we make a first step in analyzing particle swarm optimization in different versions (including pre-trained and retrained neural networks as surrogates) for its energy profile (for both processor and memory), plus a further study on the surrogate accuracy to properly drive the search towards an acceptable solution. Our conclusions shed new light on this topic and could be understood as the first step towards a methodology for assessing surrogate-assisted algorithms not only accounting for time or numerical efficiency but also for energy and surrogate accuracy for a better, more holistic characterization of optimization and learning techniques.
\end{abstract}


\section{Introduction}
Solving a problem in a real-world settlement is a really complex task. There are many considerations on the problem modeling, the interaction with final users, and the used techniques. Most ongoing research on the techniques to successfully solve a problem focuses in making them efficient and accurate~\cite{OSABA2021100888}. Efficiency usually comes down to measuring the running time of the solver~\cite{alba2002improving, ABDELHAFEZ2020100692}, with just not much attention to other computational issues, e.g., memory consumption~\cite{khalfi2023metaheuristics}. As to the accuracy of the solver, a big amount of works use the distance to the known optimum solution (or an acceptable reference solution cost) to measure how precise the technique is~\cite{He2016Convergence}. Even if many different approaches exist to these base ideas~\cite{Halim2021}, what really becomes apparent is the need for a further, more comprehensive analysis of the used techniques, since important features are usually not addressed.

One of these important features of a solver rarely addressed is its energy profile. Though some studies exist on this~\cite{Abdelhafez2019,Jamil2022}, the energy consumption of an algorithm has many implications in how likely it is to use the technique in practice, what is the cost of (super-)computing, and what is its impact on Green Computing~\cite{Paul2023Green}. Energy-aware and low consuming techniques and what modern science and society now require from artificial intelligence (AI).

Amongst the many ways of improving a basic solver (e.g., a metaheuristic) to make it attractive and competitive to solve a real problem, we can find the use of surrogates. Since the fitness (cost or benefit) function of a metaheuristic usually encodes a simulation of a whole real system, it is very time and memory-consuming to run algorithms using this function thousands or millions of times. There, surrogates like neural networks (NN) \cite{iima2023genetic} become a potential solution, acting as a kind of light proxy between a tentative solution created by the algorithm and the computation of its associated fitness, and guiding the iterative search toward an acceptable solution.

However, most studies on surrogates just go for the reductions in time that they provide~\cite{Li2022PHEV}, in some cases not even considering the overhead of their initial training. Usually, if the surrogate takes the algorithm to a satisfactory solution in a reduced time, it is considered a good idea~\cite{TONG2021414}. However, the natural need for advances in research pushes us to ask more precise questions to characterize a better design of surrogate-assisted metaheuristics. Here, we face the following ones:

RQ1: How much energy can the surrogate itself consume? How does this energy compare to the one of the real fitness function and the whole algorithm solving the problem?

RQ2: If we now have energy profiles of the original fitness function, the surrogate model, and the algorithm using either of them, how are we now judging what is an accurate and efficient algorithm?

RQ3: Do we need a perfectly trained NN for a successful algorithm, or is quick training ---low overhead--- enough for it to be useful?

RQ4: Is it better to pre-train the surrogate and then always use it as a substitute for the fitness function in the metaheuristic, or is it preferable to do a minimum initial training and start using it with later iterative training steps during the search?

These questions point to a more global methodology for surrogate-assisted algorithms, where not only obvious metrics like time, number of evaluations, or final success of the solver are included, but also the energy profile of all the components and the numerical accuracy of the surrogate model are taken into account. By answering the previous questions, we will foster new thinking on when and how surrogates are useful, and we will provide some initial data for future comparisons.

In this paper, we limit ourselves to a given search template (a canonical particle swarm optimization; PSO) and one type of surrogate (a NN) to solve a real mobility problem in smart cities (optimal tuning of the times in red of traffic lights for efficient road traffic). Even if these choices was made to allow a focused short paper, some generalizations can be made to other techniques, problems, and surrogate usage.

This article first discusses some relevant related work (Section II), then a brief description of the solved problem (Section III), to later present the family of algorithms evaluated (Section IV) in a further experimental setting (Section V). Finally, numerical results (Sections VI and VII) and fresh conclusions from the study are drafted (Section VIII).

\section{Related Work}
\subsection{Energy Consumption of Search Algorithms}


Research on energy efficiency and green computing in AI, including machine learning, has gained attention~\cite{Verdecchia2023}. Recent studies have assessed the energy efficiency of evolutionary algorithms and metaheuristics. Abdelhafez et al.~\cite{Abdelhafez2019} conducted a component-based study on the energy consumption of sequential and parallel genetic algorithms. Other works~\cite{Jamil2022} explored the energy consumption of nature-inspired optimization algorithms like genetic algorithms (GA), particle swarm optimization (PSO), differential evolution (DE), and artificial bee colony (ABC). D\'{i}az-\'{A}lvarez et al.~\cite{Diaz2022} investigated the impact of population size on the energy consumption of genetic programming (GP). Fern\'{a}ndez de Vega et al.~\cite{Fernandez2020} focused on parameters, including population size, affecting the energy efficiency of GAs. Despite these efforts, more research is needed on the energy efficiency of surrogate-assisted evolutionary algorithms.

\subsection{Surrogate Models and their Efficiency/Accuracy}


Efficiency studies on surrogate-assisted evolutionary algorithms (SAEAs)~\cite{JIN201161,HE2023119495} mainly concentrate on reducing the time or number of expensive evaluations. Works like~\cite{Cui2022SAEO} typically experimented with a limited number of evaluations, around hundreds. Addressing the time efficiency, Wei et al.~\cite{Wei2023eToSA-DE} introduced the two-stage surrogate-assisted differential evolution (eToSA-DE) algorithm, tailored for handling inefficient surrogate training with an increasing number of constraints. Notably, eToSA-DE demonstrated shorter execution times in problems with many constraints.

However, the literature on energy efficiency for surrogates is limited. In one instance, Li et al.~\cite{Li2022PHEV} proposed the surrogate-assisted strength Pareto evolutionary algorithm (SSPEA) for multiobjective optimization in energy management systems for plug-in hybrid vehicles (PHEVs). Results indicated that SSPEA saves over 44.6\% energy during the R\&D process compared to methods using only physical information.

\section{Traffic Light Scheduling}
Traffic lights play a crucial role in managing traffic flow in cities. They are placed at intersections, pedestrian crossings, and other locations to control traffic and avoid accidents. At each intersection, the traffic lights are synchronized to work in a valid sequence of phases. Each phase includes a combination of colored lights and allows vehicles to use the roadway for a certain amount of time. The assignment of the time span for each phase in the sequence of all intersections in an urban area is called a traffic light plan. Finding an optimal traffic light plan is essential to minimize the number of red light stops, thereby reducing travel time for vehicles.

The task at hand involves optimizing multiple objectives for a specific area within a particular time frame. These objectives may include maximizing the number of vehicles that reach their destination within a given period, increasing the average speed of all vehicles in the area, reducing the waiting time of vehicles at intersections, and minimizing the length of waiting queues. To represent this problem mathematically, we adopt the approach used by \cite{GARCIANIETO2012274}. This is a multi-objective problem that aims to maximize the number of vehicles arriving at their destination ($NV_D$) and minimize the number of vehicles that do not reach their destination ($NV_{ND}$) during the simulation time ($T_S$). Additionally, the problem seeks to minimize the total travel time ($TT_v$) of all vehicles from the start of the simulation to their destination and the time spent stopping and waiting by all vehicles ($TT_{EP}$). Furthermore, we maximize the ratio $P$ of green and red colors in each phase state of all intersections using the following equation:

\begin{equation}\label{eq: prob1}
P =\sum_{i=0}^{inter}\sum_{j=0}^{fs}d_{i,j}\frac{g_{i,j}}{r_{i,j}},
\end{equation}
Here, $inter$ represents the number of intersections, $fs$ represents the number of phases in each intersection, and $g_{i,j}$ and $r_{i,j}$ denote the number of green and red signal colors at intersection $i$ and phase state $j$. $d_{i,j}$ represents the duration of these signals. Lastly, all the objectives are combined into a single objective as follows:

\begin{equation}\label{eq:prob2}
F=\frac{TT_{v} + TT_{EP} + NV_{ND} \times T_{S}}{NV_{D}^{2}+P},
\end{equation}

The end goal is to minimize Eq.~\eqref{eq:prob2}. A possible solution to this problem is a vector of positive integers where each integer represents the duration of each semaphore phase of each intersection in seconds. To evaluate the effectiveness of each potential solution, we need to analyze its impact on the traffic flow of the city under consideration. To achieve this, we use SUMO (Simulation of Urban MObility) to simulate the movement of vehicles in detail, and collect the necessary data for calculating the quality of each potential solution according to Eq.~\eqref{eq:prob2}. While the results of the simulator are highly accurate and closely reflect reality, the process is time-consuming, taking several seconds to tens of minutes for each potential solution, depending on the size of the simulated urban area. This cost and time-intensive process clearly highlights the need to explore ways to speed it up, such as using surrogate systems.

\section{Surrogate-Assisted Particle Swarm Optimization}
\begin{algorithm}[tb]
    \caption{A pseudo-code of NN assisted PSO.  Black: PSO without the NN surrogate, Black+\textcolor{red}{Red}: PSO with the NN surrogate pre-training (SAPSO-p), Black+\textcolor{red}{Red}+\textcolor{blue}{Blue}: PSO with the NN surrogate retraining (SAPSO-r).}
    \label{alg:sapso}
    \begin{algorithmic}[1]
    \State Randomly generate $N$ solutions as the initial swarm $P_0$
    \State Evaluate all solutions using an actual fitness function $f$
    \State Set the global and personal best particles
    \State $FE=N, g=0$
    \color{red}
    \State Store all solutions in the dataset $D$
    \State $trained=false$
    \State $retraining=false$
    \color{black}
    \color{blue}
    \State $retraining=true$
    \color{black}
    \While{$FE<MaxFE$}
        \color{red}
        \If{$|D|\ge N_{t} \land (\lnot trained \lor retraining$)}
            \State Train NN surrogate model $\hat{f}$ from $D$
            \State $trained=true$
        \EndIf
        \color{black}
        \For{each particle $\bm{x}_g^i$}
            \State Update velocity $\bm{v}_{g+1}^i$ by Eq.~\eqref{eq:update_velocity}
            \State Update position $\bm{x}_{g+1}^i$ by Eq.~\eqref{eq:update_position}
            \color{red}
            \If{$trained$}
                \State Predict fitness value of $\bm{x}_{g+1}^i$ using $\hat{f}$
            \Else
            \color{black}
                \State Evaluate fitness value of $\bm{x}_{g+1}^i$ using $f$
                \color{red}
                \State Store the evaluated particle in the dataset $D$
            \EndIf
            \color{black}
            \State Update the personal best particle
            \State $FE = FE + 1$
        \EndFor
        \color{blue}
        \If{$trained$}
            \State Choose the top $N_r$ particles according to $\hat{f}$
            \State Evaluate fitness of the chosen particles using $f$
            \State Store the evaluated particles in the dataset $D$
        \EndIf
        \color{black}
        \State Update the global best particle
        \State $g=g+1$
    \EndWhile
    \color{red}
    \If{$\lnot retraining$}
        \State Evaluate the best solution in $P_t$ using $f$
    \EndIf
    \color{black}
    \end{algorithmic}
\end{algorithm}
Algorithm~\ref{alg:sapso} outlines the pseudo-code of the PSO algorithm used in this study, referring to the work~\cite{GARCIANIETO2012274} for traffic light scheduling problem. The PSO algorithm without the NN surrogate model, denoted as ``PSO'', is depicted in the black text (lines 1--4, 9, 14--16, 20, 23--25, and 31--33). On the other hand, the PSO algorithm with NN surrogate pre-training, labeled as ``SAPSO-p'', is shown in the black and red text, including additional lines 5--7, 10--13, 17--19, 21--22, and 34--36. Additionally, the PSO algorithm with NN surrogate retraining, termed ``SAPSO-r'', is presented in black, red, and blue text, featuring additional lines 8 and 26--30).

Initially, $N$ initial solutions are randomly generated and evaluated using an actual fitness function (see Section~\ref{sec:prob-desc}).
Subsequently, the algorithm iterates, updating the velocities and positions of particles. The position of the $i$-th particle $\bm{x}_{g}^i$ and its velocity $\bm{v}_g^i$ are updated with the following equations:
\begin{equation}
    \bm{v}_{g+1}^i\leftarrow w\cdot \bm{v}_g^i + \phi_1\cdot U(0, 1)\cdot (\bm{p}_g^i-\bm{x}_g^i)+\phi_2\cdot U(0, 1)\cdot(\bm{b}_g-\bm{x}_g^i)
    \label{eq:update_velocity}
\end{equation}
\begin{equation}
    \bm{x}_{g+1}^i\leftarrow \bm{x}_{g}^i + \bm{v}_{g+1}^i
    \label{eq:update_position}
\end{equation}
where $\bm{p}_g^i$ represents the personal best particle discovered by the $i$-th particle, $\bm{b}_g^i$ denotes the global best position found among all particles. The inertia weight $w$ regulates the momentum in the velocity update equation. Coefficients $\phi_1$ and $\phi_2$ control the influence of personal and global best particles, respectively. $U(0, 1$) corresponds to a uniform random number ranging from 0 to 1.

To accommodate integer numbers for the traffic light scheduling problem, modifications have been made to the velocity calculation using floor and ceiling functions as follows:
\begin{equation}
    v_{g+1}^i=\begin{cases}
        \lfloor v_{g+1/2}^i\rfloor&\text{if}\:U(0, 1)^i\le \lambda\\
        \lceil v_{g+1/2}^i\rceil&\text{otherwise}
    \end{cases}
    \label{eq:velocity_calculation}
\end{equation}
where $v_{g+1/2}^i$ indicates the intermediate velocity value obtained by Eq.~\eqref{eq:update_velocity}, while $\lfloor\cdot\rfloor$ and $\lceil\cdot\rceil$ indicate the floor and ceiling functions, respectively. The parameter $\lambda$ determines the probability of performing the floor or ceiling functions.

In addition, the linear inertia weight change is used through the optimization process. Concretely, the inertia weight $w$ is updated by the following equation:
\begin{equation}
    w\leftarrow w_{max}-\frac{(w_{max}-w_{min})\cdot g}{g_{total}}
    \label{eq:weight_schedule}
\end{equation}
where $w_{max}$ and $w_{min}$ are the maximum (initial) and minimum (final) weights, while $g_{total}$ is the maximum number of generations. 

During optimization, SAPSO-p and SAPSO-r store the evaluated particles in the dataset $D$ (line 5). When the dataset size $D$ exceeds the number of initial training data $N_t$, the NN surrogate is trained and utilized for fitness value predictions (lines 10--13). In SAPSO-p, the surrogate model is trained once when $N_t$ training data is obtained and is used throughout the optimization. To obtain the final result, SAPSO-p performs a single actual evaluation on the solution with the best predicted value from the final population (lines 34--36), since it only utilizes predicted values during the optimization process after training the NN surrogate. Conversely, in SAPSO-r, the top $N_r (N_r<N)$ particles with the highest predicted values undergo evaluation with an actual fitness function every generation, and the results are stored in the dataset $D$ for surrogate model retraining (lines 26--30).

\section{Experimental Settings}

\subsection{Problem Instance Description\label{sec:prob-desc}}

To study the behavior of our proposals, SAPSO variants, we utilized a problem instance situated in the large metropolitan area of M\'{a}laga. The scenario was created by extracting information from real digital maps and consists of 190 traffic lights, covering an area of $2.55$ \si{\kilo\metre\squared}. 
In the simulation, we considered 1200 vehicles, each starting from a specific point and traveling to a destination. The speed limit for the vehicles was set at 50 \si{\kilo\metre/\hour}, which is the typical maximum speed in urban areas. The simulation time was set at 2,200 \si{\second}, allowing us to collect realistic traffic information without taking an unreasonable amount of time. 
The simulation was executed using the traffic simulator SUMO release 0.32.0 for Linux.

\subsection{Parameter Settings}
\begin{table}[tb]
\centering
\caption{Parameter settings of SAPSO}
\label{tb:param_ga}
\begin{tabular}{lr}
\toprule
Parameter&Value\\
\midrule
Swarm size ($N$)&100\\
The maximum number of evaluations ($MaxFE$)&30000\\
Local coefficient ($\phi_1$)&2.05\\
Global coefficient ($\phi_2$)&2.05\\
Velocity truncation factor ($\lambda$)&0.5\\
Maximum inertia weight ($w_{max}$)&0.5\\
Minimum inertia weight ($w_{min}$)&0.1\\
The number of initial samples ($N_t$) (SAPSO)&100, 8192\\
The number of re-evaluations ($N_r$) (SAPSO-r)&10\\
\bottomrule
\end{tabular}
\end{table}

Table~\ref{tb:param_ga} outlines the parameter settings of PSO and the SAPSO variants. The swarm size (number of particles) is 100. Local and global coefficients ($\phi_1$ and $\phi_2$ in Eq.~\eqref{eq:update_velocity}) are both assigned a value of $2.05$. The velocity truncation factor $\lambda$ is set to 0.5, indicating an equal selection of ceiling and floor functions in Eq.~\eqref{eq:velocity_calculation} are equally selected. The minimum and maximum inertia weights in Eq.~\eqref{eq:weight_schedule} are 0.1 and 0.5, respectively. These parameter choices align with the guidelines from the previous study~\cite{GARCIANIETO2012274}. 

For both SAPSO-p and SAPSO-r, the number of the initial training dataset is set to $N_t=\{100, 8200\}$. The small dataset size $N_t=100$ is a minimum size, where the initial training uses the randomly generated initial swarm. The large dataset size $N_t=8200$ is adopted as the closest value to the training data set size of 8192, which provided sufficient NN accuracy in the initial experiments. Hereafter, the SAPSO algorithms with a small dataset size are denoted as SAPSO-ps and SAPSO-rs, while those with a large dataset size are denoted as SAPSO-pl and SAPSO-rl. For SAPSO-r, the number of re-evaluations $N_r$ is set to 10.

\begin{table}[tb]
\centering
\caption{Parameter settings of the NN surrogate model}
\label{tb:param_ann}
\begin{tabular}{lr}
\toprule
Parameter&Value\\
\midrule
Number of input&190\\
Number of hidden layers&2\\
Number of hidden neurons&285-190\\
Activation function&ReLU\\
Optimizer&Adam\\ 
Number of training epochs& 100\\
Batch size&32\\
Learning rate&$10^{-4}$\\
\bottomrule
\end{tabular}
\end{table}
Table~\ref{tb:param_ann} presents the parameter settings of the NN surrogate model. The number of inputs is 190, matching the dimension of the traffic light scheduling problem. The NN consists of two hidden layers, with 285 and 190 hidden neurons each. 
The activation function of the hidden neuron is ReLU. During NN training, the Adam optimizer is employed for 100 epochs, using a batch size of 32 and a learning rate of $10^{-4}$.

\subsection{System Specification}
The computer system operates on Ubuntu 18.04. It embeds an Intel(R) Xeon(R) CPU E5-1650 v2 running at a speed of 3.50GHz and 16GB of DRAM memory.
The algorithm is implemented in Python 3.7.5. Tensorflow 2.11.0\footnote{\url{https://www.tensorflow.org/}} and Keras 2.11.0\footnote{\url{https://keras.io/}} were utilized for implementing the NN surrogate. 

\subsection{How to Measure Energy}
This study employed a Python package called pyRAPL\footnote{\url{https://pyrapl.readthedocs.io/en/latest/}}, which is based on the Intel Running Average Power Limit (Intel RAPL) technology~\cite{David2010}, to measure the energy consumption of the algorithms. RAPL and its derivatives have demonstrated high reliability in previous research~\cite{Abdelhafez2019,Diaz2022,Ferro2023,anthony2020carbontracker}. The pyRAPL package enables separate measurements for CPU and memory (DRAM) energy consumption.

For the experiments, energy consumption and execution time were measured for each component of algortihms. Specifically, the measurements were divided into initialization, solution updates, solution evaluations, training of NN, and prediction using NN, creating a detailed profile. 

\section{Initial Understanding}
This section analyzes the energy profile of solution evaluation and surrogate model to understand the experimental results better.

\subsection{Experimental Procedure}
The first analysis assesses the average energy consumption and execution time during solution evaluation using the actual fitness function, involving expensive calls to SUMO. The experiment begins by generating a random solution and then evaluating its actual fitness, recording the associated energy consumption and execution time. This process is iterated $10^6$ times, and the average and standard deviation of the recorded values are computed.

The second analysis measures the average energy consumption and execution time throughout both the training and the prediction phases of an NN. The training iterations are repeated 100 times, with $N_t$ training data randomly selected from the archive obtained in the first analysis. During each training iteration, energy consumption and execution time are measured while training the NN with the selected data. Subsequently, each trained NN undergoes a testing loop where, for each testing phase, 100 test data points randomly selected from the archive are used to predict their fitness, and energy consumption and execution time are recorded. This testing phase is repeated 100 times for each trained NN. The average and standard deviation of the energy consumption and execution time during both, the training and testing phase, are then calculated. Furthermore, the mean absolute percentage error (MAPE) and the coefficient of determination ($R^2$) are computed based on the actual and predicted fitness pairs, MAPE and $R^2$ are calculated as follows:
\begin{align}
    MAPE&=\frac{1}{n}\sum_{i=1}^n\left|\frac{\hat{y}_i-y_i}{y_i}\right|\times 100,\\
    R^2&=1-\frac{\sum_{i=1}^n \left(y_i-\hat{y}_i\right)^2}{\sum_{i=1}^n \left(y_i-\bar{y}\right)^2},
\end{align}
where $y_i$ and $\hat{y}_i$ represent the actual and predicted fitness values, while $\bar{y}$ denotes the mean of the actual fitness values. The variable $n$ indicates the number of testing data points. A smaller MAPE value indicates a smaller difference between actual and predicted fitness values, with MAPE approaching 0 indicating a closer match. Conversely, $R^2$ approaching 1 indicates minimal relative residuals, signaling a robust fit between actual and predicted values.

\subsection{What is the Cost of One Single Evaluation of the Fitness Function?}
Initially, we performed an analysis of the energy consumption and execution time related to the evaluation of solutions using the actual fitness function. 

Table~\ref{tb:eval_init_data} provides the average and standard deviation of energy consumption and execution time during the actual evaluation. On average, a single solution evaluation leads to 179.24~\si{\joule} of CPU energy consumption, 9.26~\si{\joule} of DRAM energy consumption, and a time requirement of 3.54~\si{\second}. Since fitness functions are called millions of times in many solvers, needing more than 3s for each call is a problem, thus surrogates are a solution. These values can also be used in as future reference for studies on energy as a base for comparison.
\begin{table}[b]
\centering
\caption{Average and standard deviation of energy consumption [\si{\joule}] and execution time [\si{\second}] of one solution evaluation}
\label{tb:eval_init_data}
\begin{tabular}{lrr}
\toprule
Target & Ave. & Stdv. \\
\midrule
CPU energy consumption [\si{\joule}]&179.24 &(21.90)\\
DRAM energy consumption [\si{\joule}]& 9.26 &(1.41)\\
Execution time [\si{\second}]& 3.54 &(0.43)\\
\bottomrule
\end{tabular}
\end{table}

\subsection{Energy, Time, and Accuracy of the Surrogate}
\begin{table*}[tb]
    \centering
    \caption{Averages of energy consumption [\si{\joule}] and execution time [\si{\second}] of the NN training and testing, with standard deviations in parentheses, and the prediction accuracy}
    \label{tb:training_init_data}
    \begin{tabular}{ll*{7}{r}}
    \toprule
    &&\multicolumn{7}{c}{Dataset size}\\
    \cmidrule(lr){3-9}
    &Target&128&256&512&1024&2048&4096&8192\\
    \midrule
\multirow{3}{*}{Training}&CPU energy [\si{\joule}]   &  71.14 (1.78) &  104.66 (0.48) &  171.45 (1.13) &  304.57 (2.38) &  571.39 (4.20) &  1,078.80 (5.17) &  2,140.74 (6.38) \\
&DRAM energy [\si{\joule}]   &   1.30 (0.36) &    2.65 (0.44) &    5.64 (0.67) &   10.82 (1.41) &   18.44 (2.61) &    96.32 (2.32) &   197.57 (1.57) \\
&Execution time [\si{\second}]         &   1.90 (0.04) &    2.89 (0.03) &    4.98 (0.05) &    9.20 (0.09) &   17.67 (0.14) &    33.46 (0.26) &    67.65 (0.32) \\
\midrule
\multirow{3}{*}{Testing}&CPU energy [\si{\joule}]  &  1.98 (0.41) &  1.98 (0.45) &  1.99 (0.44) &  1.99 (0.41) &  1.99 (0.36) &  1.96 (0.39) &  1.97 (0.40) \\
&DRAM energy [\si{\joule}] &  0.03 (0.03) &  0.04 (0.03) &  0.05 (0.03) &  0.06 (0.03) &  0.05 (0.03) &  0.15 (0.05) &  0.15 (0.05) \\
&Execution time [\si{\second}]         &  0.05 (0.01) &  0.05 (0.01) &  0.05 (0.01) &  0.05 (0.01) &  0.05 (0.01) &  0.05 (0.01) &  0.05 (0.01) \\
\midrule
\multirow{2}{*}{Accuracy}&MAPE [\%]        &        33.59 &        31.47 &        29.15 &        27.40 &        25.73 &        23.69 &        21.55 \\
&$R^2$                             &        -0.06 &         0.05 &         0.15 &         0.22 &         0.29 &         0.40 &         0.50 \\
    \bottomrule
    \end{tabular}
\end{table*}
Next, we analyzed energy consumption, execution time, and prediction error for the NN surrogate model. The NN was trained across various dataset sizes, \{128, 256, 512, 1024, 2048, 4096, and 8192\}. Table~\ref{tb:training_init_data} summarizes the average and standard deviation of energy consumption, execution time, and NN training and prediction accuracy.

The CPU energy consumption increases linearly but moderately with dataset size. Despite a 64 times increase in dataset size from 128 to 8192, CPU energy consumption only grows approximately 30 times. Execution time follows a similar trend, where a 64 times increase in dataset size corresponds to a 35.6 times increase in execution time. In contrast, DRAM energy consumption rises sharply for dataset sizes of 4096 and 8192, reaching 74 and 151 times, exceeding the corresponding dataset size increase.

Energy consumption and execution time during prediction remain constant regardless of the training dataset size. However, prediction accuracy improves with larger training datasets. MAPE averages 33\% for a training dataset size of 128, reducing to 21\% for a dataset size of 8192, representing an improvement of over 10\%. For $R^2$, smaller training dataset sizes approach 0 (or negative), indicating larger residuals. Larger training datasets yield improved predictions, with $R^2=0.50$, signifying a better fit.

Comparing energy consumption between actual evaluation (Table~\ref{tb:eval_init_data}) and prediction (Table~\ref{tb:training_init_data}), training with a dataset size of 512 shows similar energy consumption and execution time to actual solution evaluation. However, training with a larger dataset requires significantly more energy consumption and execution time, particularly 12 times higher energy consumption and 19 times longer execution time for a dataset size of 8192. Conversely, NN prediction is about 90 times more energy-efficient and approximately 68 times faster in terms of execution time. Thus, even when trained with a large dataset, substituting solution evaluation with NN not only reduces execution time but also minimizes energy consumption.

\section{Experimental Results}


In this section, we conduct three experiments to analyze the energy consumption, execution time, and search performance of SAPSO. The experiments are as follows:

\begin{itemize}
\item{\textbf{Experiment 1:} Study the physical performance ---energy and time--- of algorithms (to answer RQ1 and RQ2).}
\item{\textbf{Experiment 2:} Study the numerical performance ---accuracy--- of algorithms (to answer RQ3 and RQ4).}
\item{\textbf{Experiment 3:} Study the accuracy of the surrogate versus the real fitness (to answer RQ3 and RQ4).}
\end{itemize}

The SAPSO variants were executed with 20 independent runs each. PSO was executed 11 runs due to the time limitation, with the additional nine runs on a computer with different specifications. These additional runs were used to obtain the search performance only and not to analyze energy and time.

\subsection{Experiment 1: Energy Consumption and Time}
\begin{table*}
\centering
\caption{Average energy consumption and execution time for each optimization component and standard deviation}
\label{tab:measurement_summary}
\scalebox{0.8}{
\begin{tabular}{llrrrrrrrrrrrr}
\toprule
& & \multicolumn{2}{c}{Evaluation} & \multicolumn{2}{c}{Training} & \multicolumn{2}{c}{Prediction} & \multicolumn{2}{c}{Initialization} & \multicolumn{2}{c}{Update}&\multicolumn{2}{c}{Total} \\
\cmidrule(lr){3-4}\cmidrule(lr){5-6}\cmidrule(lr){7-8}\cmidrule(lr){9-10}\cmidrule(lr){11-12}\cmidrule(lr){13-14}
& Target& Ave. & Stdv. & Ave. & Stdv. & Ave. & Stdv. & Ave. & Stdv. & Ave. & Stdv.  & Ave. & Stdv.\\
\midrule
\multirow{3}{*}{PSO}& CPU Energy [\si{\joule}] & 3,592,542.58 & (88,126.38) & NA & NA & NA & NA & 0.75 &(0.03) & 454.44& (15.90)& 3,592,997.77 & (88,131.71)\\
& DRAM Energy [\si{\joule}] & 254,015.06 & (49,583.03) & NA & NA & NA & NA & 0.05 &(0.01) & 44.76& (7.58)& 254,059.87 & (49,590.53)\\
& Execution time [\si{\second}] & 70,834.13 & (1,672.66) & NA & NA & NA & NA & 0.01 &(0.00) & 9.67& (0.23)& 70,843.81 & (1,672.71)\\
\midrule
\multirow{3}{*}{SAPSO-ps}& CPU Energy [\si{\joule}] & 17,591.22 & (201.42) & 173.36& (1.70) & 58,377.47& (253.27) & 0.75 &(0.03) & 276.64& (6.36)& 76,419.44 & (347.20)\\
& DRAM Energy [\si{\joule}] & 1,545.41 & (30.89) & 14.10& (0.23) & 4,368.56& (83.37) & 0.06 &(0.01) & 16.70& (0.44)& 5,944.84 & (101.87)\\
& Execution time [\si{\second}] & 352.69 & (4.00) & 4.29& (0.07) & 1,535.36& (5.67) & 0.01 &(0.00) & 4.46& (0.08)& 1,896.83 & (7.65)\\
\midrule
\multirow{3}{*}{SAPSO-pl}& CPU Energy [\si{\joule}] & 1,025,641.47 & (29,629.21) & 2,028.96& (466.98) & 43,219.66& (343.71) & 16.09 &(20.74) & 315.52& (10.61)& 1,071,221.70 & (29,735.79)\\
& DRAM Energy [\si{\joule}] & 75,705.83 & (16,662.41) & 136.46& (46.91) & 2,372.01& (791.05) & 1.42 &(1.84) & 21.35& (5.72)& 78,237.07 & (17,497.25)\\
& Execution time [\si{\second}] & 20,352.07 & (732.83) & 62.55& (1.10) & 1,119.95& (12.70) & 0.34 &(0.44) & 5.81& (0.22)& 21,540.72 & (743.16)\\
\midrule
\multirow{3}{*}{SAPSO-rs}& CPU Energy [\si{\joule}] & 392,516.02 & (8,187.23) & 159,584.36& (3,262.53) & 60,057.30& (672.87) & 1.45 &(2.08) & 290.06& (8.59)& 612,449.19 & (8,791.63)\\
& DRAM Energy [\si{\joule}] & 29,055.48 & (6,077.90) & 10,413.69& (3,147.22) & 3,378.19& (1,097.48) & 0.13 &(0.23) & 14.05& (4.35)& 42,861.54 & (10,312.86)\\
& Execution time [\si{\second}] & 7,897.83 & (190.15) & 4,506.91& (20.70) & 1,556.73& (17.14) & 0.03 &(0.04) & 4.79& (0.05)& 13,966.30 & (217.10)\\
\midrule
\multirow{3}{*}{SAPSO-rl}& CPU Energy [\si{\joule}] & 1,276,871.54 & (32,227.93) & 507,607.74& (9,338.46) & 44,150.46& (1,276.70) & 5.26 &(9.39) & 329.54& (10.35)& 1,828,964.53 & (34,264.79)\\
& DRAM Energy [\si{\joule}] & 94,474.59 & (19,235.85) & 34,321.24& (10,470.22) & 2,514.56& (817.24) & 0.54 &(1.05) & 23.24& (6.36)& 131,334.17 & (30,449.84)\\
& Execution time [\si{\second}] & 25,395.77 & (766.96) & 15,580.22& (123.81) & 1,146.12& (42.01) & 0.11 &(0.20) & 6.18& (0.48)& 42,128.39 & (860.60)\\
\bottomrule
\end{tabular}
}
\end{table*}
Table~\ref{tab:measurement_summary} presents average energy consumption and execution time for each algorithm, with metrics measured at each algorithm component and the total displayed in the ``Total’’ column. Note that PSO without the surrogate model does not involve NN training and prediction, resulting in ``NA’’.

In terms of energy consumption, PSO allocates energy mainly to solution evaluations, with negligible consumption for initialization and solution updates. Similarly, SAPSO variants exhibit minimal energy consumption for initialization and updates. However, additional energy is consumed for surrogate training and prediction. Analyzing the energy consumption for training, algorithms with pre-training (SAPSO-ps and SAPSO-pl) demonstrate lower energy consumption, particularly SAPSO-ps with the least, owing to a single pre-training. In contrast, SAPSO-rs and SAPSO-rl, involving retraining, incur higher energy consumption, with SAPSO-rs consuming less than SAPSO-rl. For prediction tasks, the SAPSO variants employing small datasets (SAPSO-ps and SAPSO-rs) consume more energy than their counterparts using larger datasets (SAPSO-pl and SAPSO-rl). This is due to their early initiation of surrogate training, leading to an increased number of NN predictions during the optimization process.

Comparing PSO and SAPSO variants, all variants consistently demonstrate reduced total energy consumption for both CPU and DRAM compared to PSO without the surrogate model. The energy consumption for velocity and position updates is minimal, with negligible differences observed. Energy consumption related to evaluation in SAPSO variants is lower in PSO. However, the use of surrogates introduces some additional energy consumption for training and predicting evaluation values, but it remains significantly lower than actual solution evaluations. Then, it can be concluded that the SAPSO variants effectively mitigate overall energy consumption.

Among SAPSO variants, SAPSO-ps exhibits the lowest energy consumption, followed by SAPSO-rs, SAPSO-pl, and SAPSO-rl. This is attributed to SAPSO-ps requiring the fewest actual solution evaluations, with NN training performed only once. SAPSO-rs, the second most energy-efficient, involves a higher number of NN training and predictions but employs small datasets for training (at most 3000 solutions), resulting in lower energy consumption. In the comparison of SAPSO-pl and SAPSO-rl, energy consumption increases for SAPSO-rl due to repeated actual evaluations and retraining of NN during algorithm execution. A similar trend is observed in terms of execution time, with SAPSO-ps being the shortest, followed by SAPSO-rs, SAPSO-pl, SAPSO-rl, and PSO.

These results indicate that SAPSOs achieve efficient algorithms not only in execution time but also in energy consumption by replacing computationally expensive actual solution evaluations with surrogates.

\subsection{Experiment 2: Accuracy of the Algorithm}
\begin{figure}[tb]
\begin{minipage}[t]{0.48\columnwidth}
    \centering
    \includegraphics[width=\textwidth]{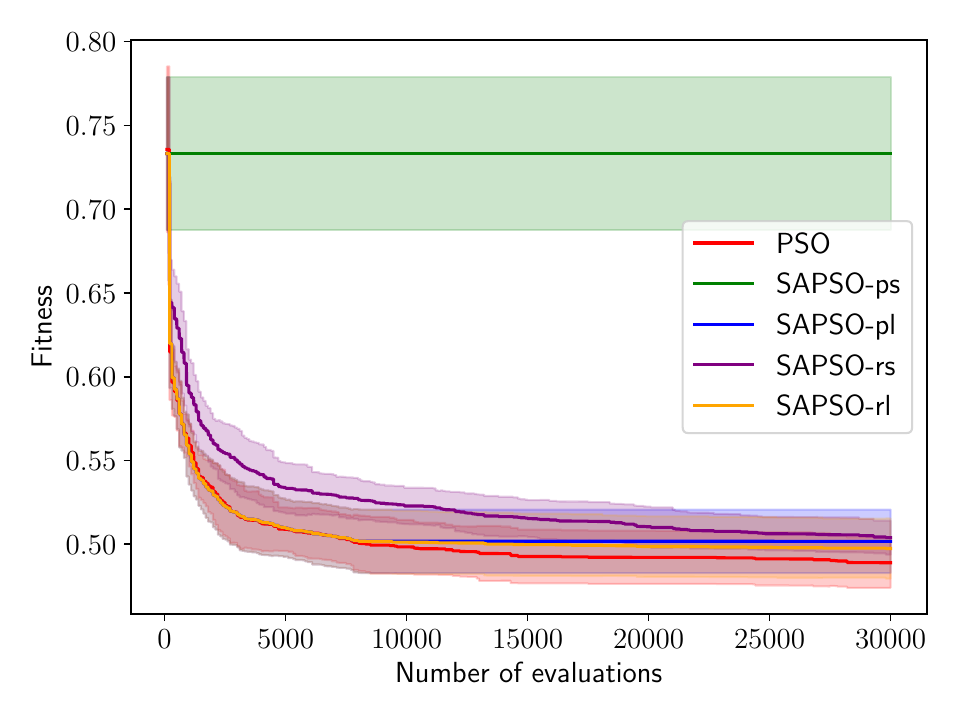}
    \caption{Transition of average fitness and standard deviation}
    \label{fig:evaluation_best_fit}
\end{minipage}
\hspace{1em}
\begin{minipage}[t]{0.48\columnwidth}
    \centering
    \includegraphics[width=\textwidth]{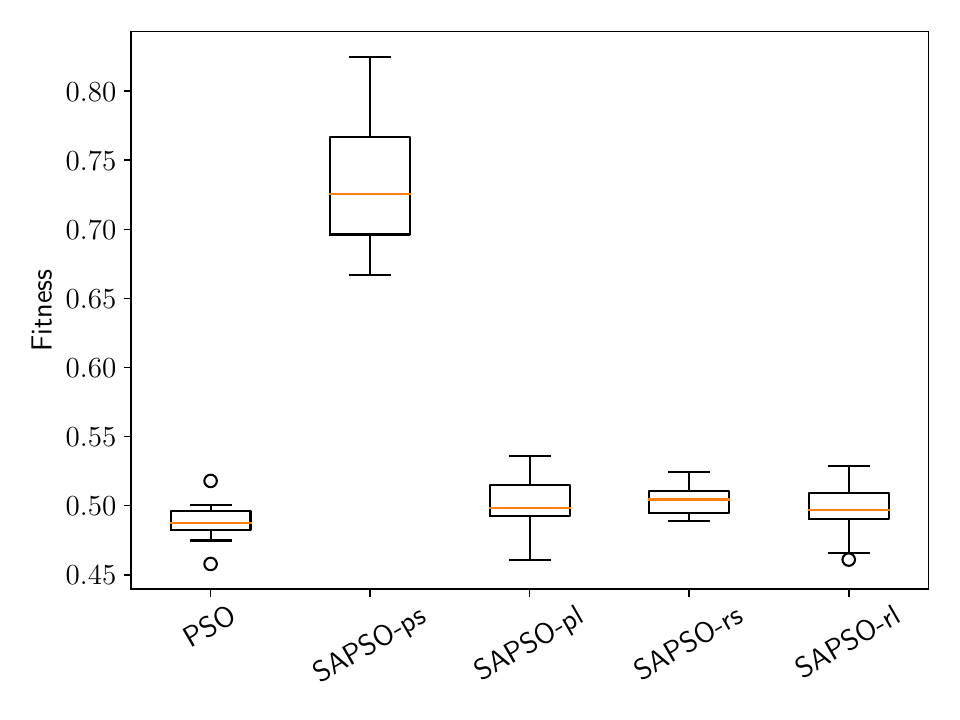}
    \caption{Boxplot of the obtained best fitness}
    \label{fig:final_fit}
\end{minipage}
\end{figure}
Fig.~\ref{fig:evaluation_best_fit} depicts the average best fitness value until reaching the maximum number of fitness evaluations. The horizontal axis shows the number of fitness evaluations, while the vertical axis shows the average actual fitness. Although the algorithm utilizing surrogates incorporates predicted evaluation values, the graphs only show values from actual evaluations.

A distinctive observation is SAPSO-ps, showing no improvement in the best fitness from the initial population. A similar trend is seen in SAPSO-pl, where the fitness value decreases until 8200 evaluations, but minimal improvement is observed after transitioning to the optimization with NN surrogate. Conversely, the other algorithms consistently reduce fitness. Specifically, SAPSOs with retraining exhibit the capability to decrease fitness even after the initial training.

Fig.~\ref{fig:final_fit} displays a boxplot illustrating the final fitness obtained with each algorithm. The horizontal axis denotes the algorithm, while the vertical axis represents the final fitness. Fig.~\ref{fig:final_fit} indicates that SAPSO-ps exhibits the worst performance. When comparing pre-training and retraining algorithms, SAPSO-pl and SAPSO-rl (algorithms utilizing large datasets) show no significant difference in the distribution of the final fitness. This suggests that, despite retraining during the optimization process, SAPSO-rl does not contribute to improving search performance. Conversely, for SAPSO-ps and SAPSO-rs (algorithms using small datasets), it is evident that retraining significantly enhances search performance. With a stable search performance, SAPSO-rs can obtain solutions comparable to SAPSO-pl, SAPSO-rl, and PSO.

\begin{figure}[tb]
\begin{minipage}[t]{0.48\columnwidth}
    \centering
    \includegraphics[width=\textwidth]{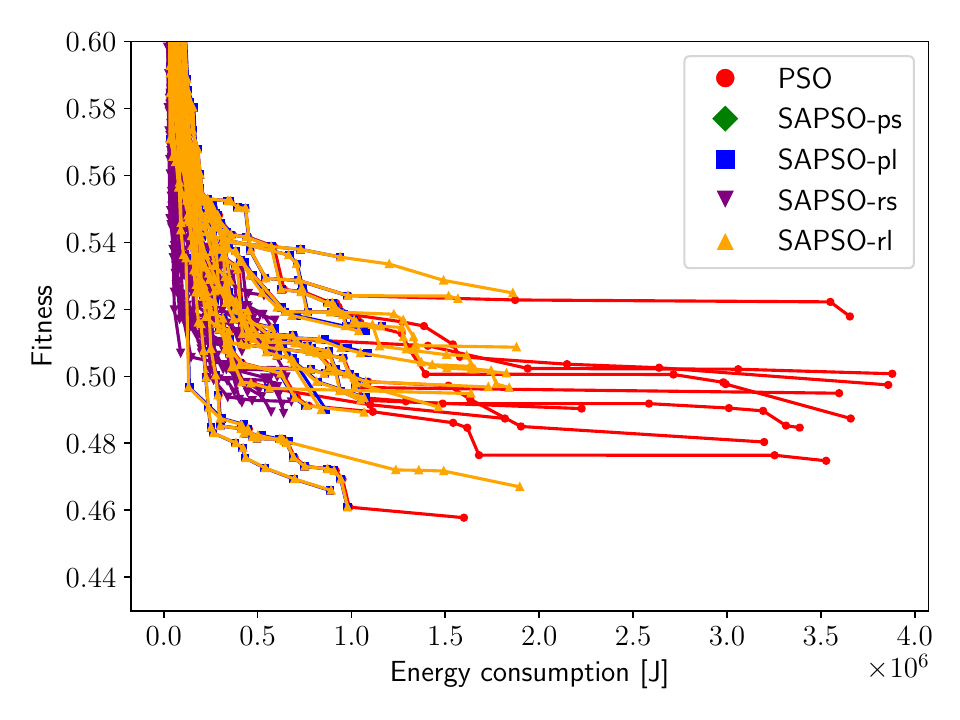}
    \caption{Obtained fitness value in relation to energy consumption [\si{\joule}]}
    \label{fig:energy_best_fit}
\end{minipage}
\hspace{1em}
\begin{minipage}[t]{0.48\columnwidth}
    \centering
    \includegraphics[width=\textwidth]{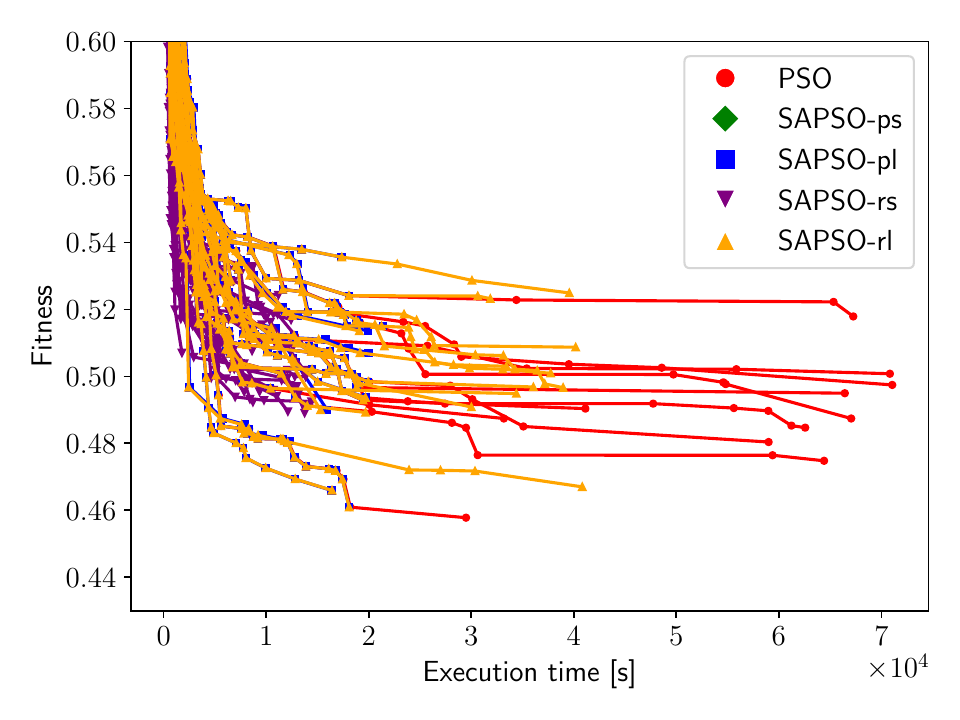}
    \caption{Obtained fitness value in relation to execution time [\si{\second}]}
    \label{fig:duration_best_fit}
\end{minipage}
\end{figure}
Figs.~\ref{fig:energy_best_fit} and \ref{fig:duration_best_fit} depict the obtained fitness value in relation to energy consumption and execution time. The horizontal axis of Fig.~\ref{fig:energy_best_fit} represents the energy consumption of the sum of CPU and DRAM, while that of Fig.~\ref{fig:duration_best_fit} represents the execution time. The vertical axis in both figures corresponds to fitness.

From these results, PSO exhibits a decrease in fitness values while consuming significant energy and execution time. Conversely, SAPSOs with retraining achieve reduced fitness with lower energy consumption and execution time. Particularly, SAPSO-rs rapidly decreases fitness and attains final results with less than 1 \si{\mega\joule} and 15000 \si{\second}.

\subsection{Experiment 3: Accuracy of the Surrogate}
\begin{figure*}[tb]
    \centering
    \begin{subfigure}{0.24\linewidth}
    \includegraphics[width=\linewidth]{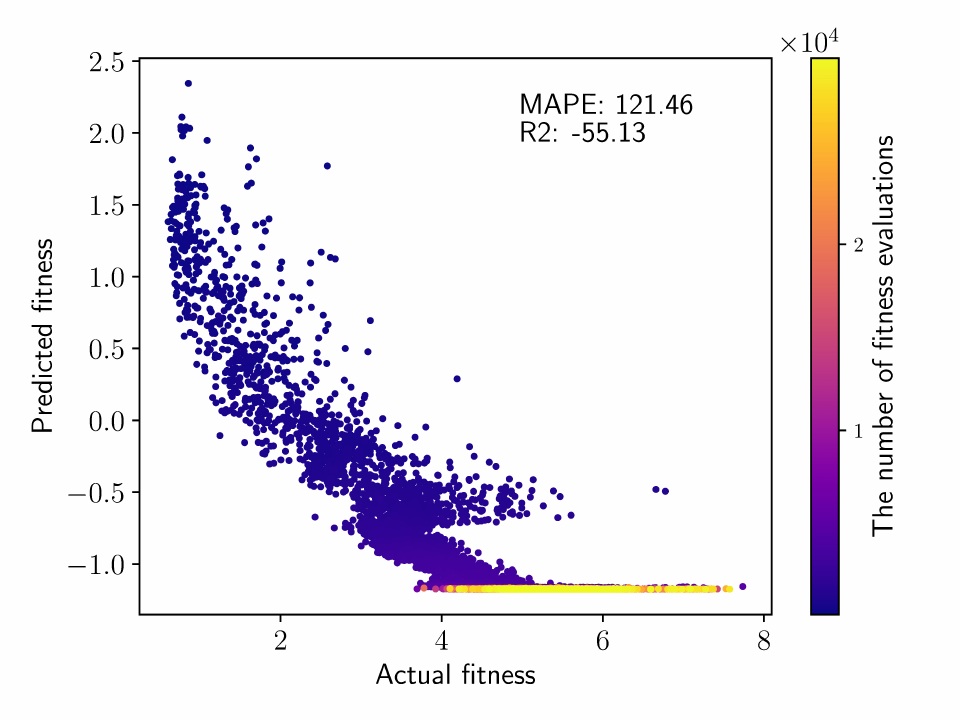}
    \caption{SAPSO-ps}
    \end{subfigure}
    \begin{subfigure}{0.24\linewidth}
    \includegraphics[width=\linewidth]{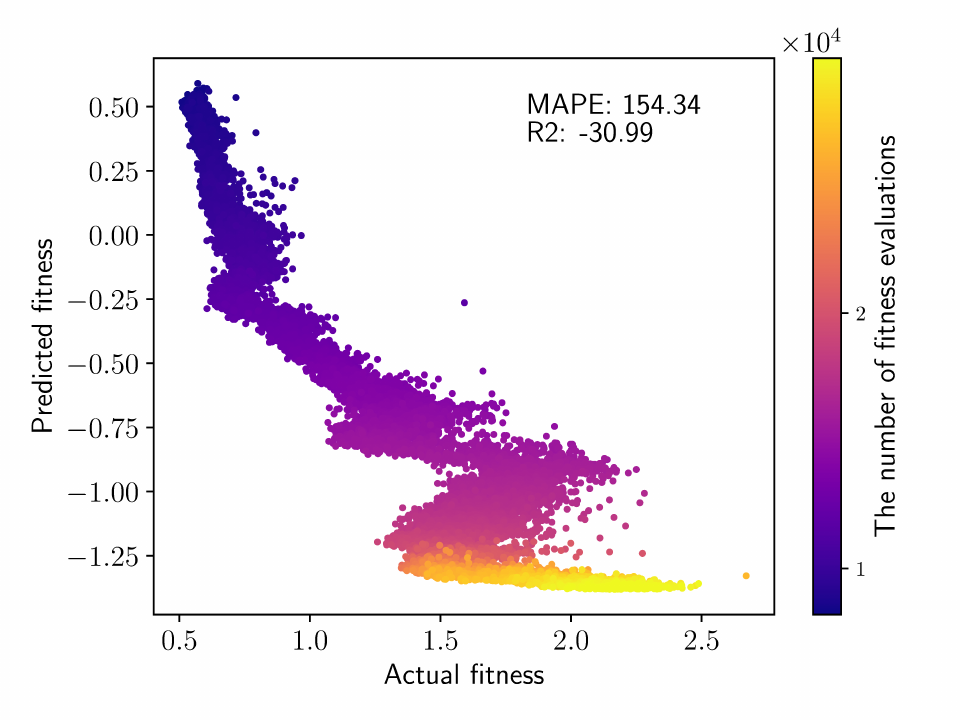}
    \caption{SAPSO-pl}
    \end{subfigure}
    \begin{subfigure}{0.24\linewidth}
    \includegraphics[width=\linewidth]{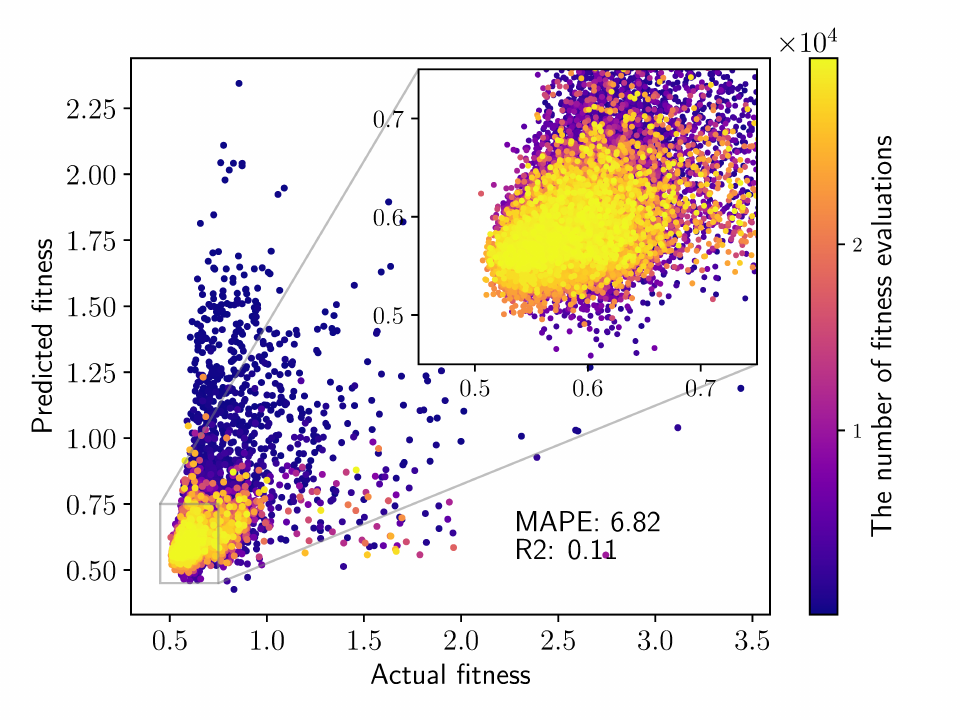}
    \caption{SAPSO-rs}
    \end{subfigure}
    \begin{subfigure}{0.24\linewidth}
    \includegraphics[width=\linewidth]{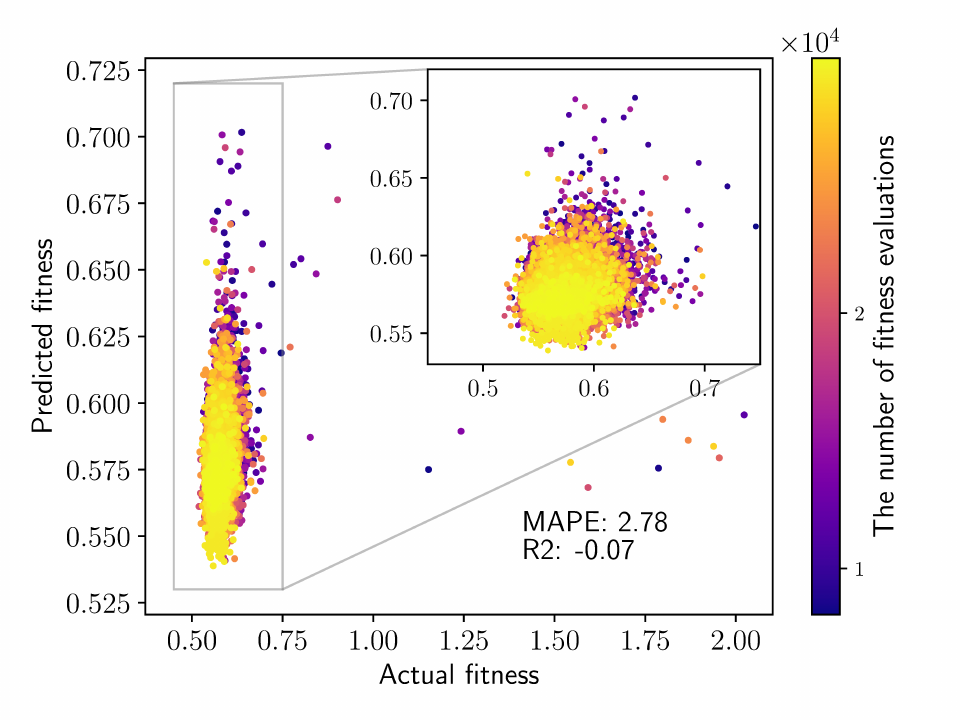}
    \caption{SAPSO-rl}
    \end{subfigure}
    \caption{Scatter plot between the actual fitness and the predicted one in a single run, when using SAPSOs}
    \label{fig:actual_predicted}
\end{figure*}
Fig.~\ref{fig:actual_predicted} illustrates the relationship between the actual and predicted fitness for the SAPSO variants. Predicted fitnesses are utilized during SAPSO execution, but actual evaluation values are independently assessed after the algorithm execution. The horizontal and vertical axes represent the actual and predicted fitness, respectively, with the color indicating the number of fitness evaluations for each point. Blue indicates the early stage, while yellow represents the final one.

Firstly, both SAPSOs employing pre-training exhibit a substantial discrepancy between the actual and predicted evaluation values, indicating the utilization of inaccurate prediction models. The error increases as the iterations progress, suggesting that optimization is advancing in regions where predicting solutions becomes challenging due to extrapolation from the initial training dataset. Although the results presented in Fig.~\ref{fig:actual_predicted} are from a single trial, similar trends have been consistently observed in other trials.

Next, the two SAPSOs with retraining can construct more accurate surrogates than SAPSOs with pre-training. Furthermore, these algorithms can accurately predict evaluation values even after advancing through iterations. This indicates that retraining allows the utilization of the surrogate in exploration while maintaining its accuracy. The accuracy of the surrogate model for SAPSO-rs was $MAPE=6.82\%$ and $R^2=0.11$, while the accuracy for SAPSO-rl was $MAPE=2.78\%$ and $R^2=-0.07$. Although both $R^2$ values are inferior to the best accuracy ones presented in Table~\ref{tb:training_init_data}, the success of the algorithms suggests that achieving a certain level of surrogate accuracy can enable efficient exploration from the perspectives of energy and time.

\section{Conclusion and Future Works}
This study aimed to investigate the energy and overall search performance of surrogate-assisted algorithms, using PSO as the base algorithm for a real (optimal traffic light orchestration) problem.

A first conclusion is that surrogate-assisted solvers are a great idea if we aim for reduced times and energy consumption. We have also learned that we can fine tune the NN for higher precision in this problem by only enlarging the data size for training. An expected higher training time, memory, and energy appear (enormous for the three larger datasets), but the later use of the trained NN remains basically constant. In fact, one estimation of the NN of a fitness value consumes 1\% of the energy of a real fitness computation, which represents a huge saving (RQ1).

Our experiments compared PSO with and without surrogates. We measured the accuracy of the obtained solutions, as well as the energy consumption and execution time. There is not a single conclusion here, but a multivariate space of results that needs a global methodology to define what is successful for the study at hand (RQ2).

The experimental results indicated that the pre-training algorithm using a low-accuracy NN leads to exploration based on inaccurate predictions, resulting in poor solutions. Even with a well-trained NN learning from a large dataset, the pre-training family of algorithms showed no improvements in its results. A real problem is prone to have difficult search landscapes and the surrogate needs a good learning of them for the used problem. We do need good trained NNs for this problem, but using a pre-trained model does not seem the best way unless we can afford long initial training times (RQ3).

In the same direction, the retraining algorithm, which repeats training during exploration, continuously enhances the accuracy of the NN throughout the search. Despite consuming much energy and time, SAPSO-rl does not contribute to improving search performance. On the other hand, the algorithm that gradually repeats training during exploration from a minimal initial dataset (SAPSO-rs) achieved comparable search accuracy and minimal energy consumption and execution time to other algorithms (RQ4). This suggests that an algorithm with small retraining cycles outperforms others in terms of accuracy, energy consumption, and execution time.

In future research, we aim to conduct a comprehensive analysis using other algorithms, such as genetic algorithms, and alternative surrogates (e.g. random forests). Additionally, we plan to explore studies involving different problem instances to assess the actual bounds to our findings.

\bibliography{references}
\bibliographystyle{IEEEtran}

\end{document}